# Improving DPLL Solver Performance with Domain-Specific Heuristics: the ASP Case

Marcello Balduccini

Intelligent Systems, KRL
Eastman Kodak Company
Rochester, NY 14650-2102 USA
marcello.balduccini@gmail.com

**Abstract** In spite of the recent improvements in the performance of the solvers based on the DPLL procedure, it is still possible for the search algorithm to focus on the wrong areas of the search space, preventing the solver from returning a solution in an acceptable amount of time. This prospect is a real concern e.g. in an industrial setting, where users typically expect consistent performance. To overcome this problem, we propose a framework that allows learning and using domain-specific heuristics in solvers based on the DPLL procedure. The learning is done off-line, on representative instances from the target domain, and the learned heuristics are then used for choice-point selection. In this paper we focus on Answer Set Programming (ASP) solvers. In our experiments, the introduction of domain-specific heuristics improved performance on hard instances by up to 3 orders of magnitude (and 2 on average), nearly completely eliminating the cases in which the solver had to be terminated because the wait for an answer had become unacceptable.

## 1 Introduction

In recent years, solvers based on the DPLL procedure [1,2] have become amazingly fast. Mostly, that is due to good heuristics that direct the search toward the most promising areas of the search space, and to learning algorithms that discover features of the search space on-the-fly. Unfortunately, when the search space is sufficiently large, it is still possible for the search algorithm to mistakenly focus on areas of the search space that contain no solutions or very few. When that happens, performance degrades substantially, even to the point that the solver may need to be terminated before returning an answer. This prospect is a real concern when one is considering using such a solver in an industrial application, in which the solver will act as part of a black-box from which users typically expect consistent performance. It should be noted that the phenomenon of performance degradation is often due to the fact that the heuristics used in choice-point selection are general-purpose, and thus may not fit well a given domain.

Various methods have been proposed in the literature to improve solver stability. A basic technique involves using parametrized general-purpose heuristics, for which the user can manually specify parameter values that are suitable for the domain of interest. An interesting step in that direction is provided by CLASPFOLIO [3], which makes use of different configurations of the CLASP solver [4]. The program in input is automatically analyzed, and the most promising configuration of CLASP is selected accordingly. A different approach consists in having the solver adapt to the problem in input at run-time by means of learning. This is the case of the clause learning and conflict learning techniques that have recently become very popular especially in SAT and ASP solvers (see e.g. [5,4]), and have brought about substantial performance improvements. The idea behind these learning techniques is to record information about the conflicts that are detected during the exploration of the search space, and to use the information to avoid descending similar branches later. Hence, the basic heuristic is still general-purpose, but it is adjusted, during execution, depending on the features of the problem in input. One drawback of this approach is that the learning is limited to the current program, and the information that has been learned in one run cannot be used in later runs of the solver.

In this paper we propose a framework, called DORS, which instead allows learning and using domain-specific heuristics in solvers based on the DPLL procedure. The results of learning are retained and can be used in later runs. In fact, the learning technique *relies* on the availability of information from multiple runs of the solver. Although here we focus on solvers for Answer Set Programming (ASP) [6,7], the DORS framework can be applied to any solver based on the DPLL procedure, including SAT and constraint solvers. Furthermore, the current framework is aimed at improving the efficiency of the computation of one model of consistent programs, but could be extended further. The learning is done off-line, on representative instances from the target domain. The particular learning technique used here is intendedly extremely simple, but already shows remarkable performance improvements. In experimental evaluation, the use of our technique improved performance on hard instances by up to 3 orders of magnitude (and 2 on average on industrial problems), nearly completely eliminating the situations in which the solver had to be terminated because the wait for an answer had become unacceptable.

This paper is organized as follows. In the next section we give some background on ASP. Next, we discuss the basic search algorithm used in most ASP solvers. Then, in Section 4, we present the DORS framework. Experimental results are discussed in Section 5. Finally, in Section 6, we draw conclusions.

## 2 Answer Set Programming

We define the syntax of the language precisely, but only give its informal semantics in order to save space. We refer the reader to [6,8] for a specification of the formal semantics. Let $\Sigma$ be a signature containing constant, function and predicate symbols. Terms and atoms are formed as usual in first-order logic. A (basic) literal is either an atom $a$ or its strong (also called classical or epistemic) negation $\neg a$. The set of literals formed from $\Sigma$ is denoted by $lit(\Sigma)$. A *rule* is a statement of the form:

$$h_1 \ \vee \ \ldots \ \vee \ h_k \leftarrow l_1, \ldots, l_m, \text{not } l_{m+1}, \ldots, \text{not } l_n$$

where $h_i$'s and $l_i$'s are ground literals and *not* is the so-called *default negation*. The intuitive meaning of the rule is that a reasoner who believes $\{l_1, \ldots, l_m\}$ and has no reason to believe $\{l_{m+1}, \ldots, l_n\}$, has to believe one of $h_i$'s. The part of the statement to the left of $\leftarrow$ is called *head*; the part to its right is called *body*. Symbol $\leftarrow$ can be omitted if no $l_i$'s are specified. Often, rules of the form $h \leftarrow$ not $h, l_1, \ldots,$ not $l_n$ are abbreviated into $\leftarrow l_1, \ldots,$ not $l_n$, and called *constraints*. The intuitive meaning of a constraint is that its body must not be satisfied. A rule containing variables is interpreted as the shorthand for the set of rules obtained by replacing the variables with all the possible ground terms (called *grounding* of the rule). A *program* is a pair $\langle \Sigma, \Pi \rangle$, where $\Sigma$ is a signature and $\Pi$ is a set of rules over $\Sigma$. We often denote programs just by the second element of the pair, and let the signature be defined implicitly. In that case, the signature of $\Pi$ is denoted by $\Sigma(\Pi)$. Finally, an *answer set* (or *model*) of a program $\Pi$ is a collection of its consequences under the answer set semantics. Because a convenient representation of alternatives is often important in the formalization of knowledge, the language of ASP has been extended with *constraint literals* [8], which are expressions of the form $m\{l_1, l_2, \ldots, l_k\}n$, where $m$ and $n$ are arithmetic expressions and $l_i$'s are basic literals as defined above. A constraint literal is satisfied whenever the number of literals that hold from $\{l_1, \ldots, l_k\}$ is between $m$ and $n$, inclusive. Using constraint literals, the choice between $p$ and $q$, under some set of conditions $\Gamma$, can be compactly encoded by the rule $1\{p, q\}1 \leftarrow \Gamma$. A rule of this form is called *choice rule*. When solving sets of problems from a given domain of interest, ASP programs are often divided into a *domain description* and a *problem instance*. Intuitively, the domain description encodes a description of the problem domain and of the solutions, while each problem instance encodes a different problem from the domain.

## 3 Search in ASP Solvers

The search algorithm used by many ASP solvers (e.g. SMODELS [9], DLV [10]) is based on the DPLL procedure [1,2]. The basic algorithm for the computation of a single answer set, which we will later refer to as *standard algorithm*, is show in Figure 1. The

```
function solve ( Π : Program, A : Set of Extended Literals )
    B := expand(Π, A);
    if (B is answer set of Π) then return B;
    if (B is not consistent or B is complete) then return ⊥;
    e := choose_literal(Π, B);
    B' := solve(Π, B ∪ {e});
    if (B' = ⊥) then B' := solve(Π, B ∪ {not(e)});
    return B';
```

**Figure 1.** Basic Search Algorithm for ASP

algorithm is based on the idea of growing a particular set of (ground) literals, often called partial answer set, until it is either shown to be an answer set of the program, or it becomes inconsistent. To achieve this, guesses have to be made as to which literals may be in the answer set. Let us now describe the algorithm more precisely. By *extended literal* we mean a literal $l$ or the expression *not* $l$, intuitively meaning that $l$ is known not to hold in the answer set (but its complement, $\bar{l}$, may or may not hold). Given an extended literal $e$, $not(e)$ denotes the expression $not\ l$ if $e = l$ and it denotes $l$ if $e = not\ l$. Algorithm $solve$ takes as input program, $\Pi$, and partial answer set, $A$, which is a set of extended literals. $A$ is initially empty. Function `expand` [9] is then used to add to the partial answer set all the literals that must hold given $\Pi$ and $A$. If the result of `expand` is an answer set of $\Pi$, the algorithm returns it (and terminates). If instead a contradiction is discovered, then the algorithm returns no model ($\bot$). In all other cases, the partial answer set is still incomplete but consistent. Then, function $choose\_literal$ selects an extended literal $e$ such that neither $e$ nor $not(e)$ occur in $B$. This is called the *choice literal* or *choice point*. The algorithm then calls itself recursively in order to find an answer set of $\Pi$ from the partial answer set $B \cup \{e\}$. If one such answer set is found, then the algorithm returns it. If instead no answer set is found, then the algorithm attempts to find an answer set of $\Pi$ that contains $B \cup \{not(e)\}$. If the attempt succeeds, the answer set is returned. Otherwise, the algorithm returns no model ($\bot$).

It is not difficult to see how the choices made by $choose\_literal$ greatly influence the number of choice points picked by the algorithm, and ultimately its performance. Consider for example the program:

$$P_1 = \begin{cases} p \leftarrow not\ q. \quad q \leftarrow not\ p. \\ \\ r. \\ \leftarrow p, r. \\ \leftarrow q, not\ s. \\ \\ u(X) \leftarrow t(X), not\ v(X). \\ v(X) \leftarrow t(X), not\ u(X). \\ \\ t(0).\ t(1).\ \ldots\ t(1000). \end{cases}$$

The program is clearly inconsistent. In fact, the first two rules force either $p$ or $q$ to hold, but the next three rules forbid $p$ and $q$ from holding. So, if the first call to $choose\_literal$ were to select e.g. $not\ p$, then the following call to `expand` would conclude that $q$ must hold, and that inconsistency follows (since $s$ is not defined by any rule and thus the body of the corresponding constraint is satisfied). The algorithm would then backtrack and select $p$. This time, `expand` would derive inconsistency from the fact that the body of the first constraint is satisfied. Hence, the algorithm would return $\bot$ (no model). However, consider what would happen if $choose\_literal$ were to select $u(0)$ instead of $not\ p$. Function `expand` would derive the consequence $not\ v(0)$ and fail to reach inconsistency. Then, the algorithm would recurse, and possibly select say $u(1)$. As before, `expand` would not detect any inconsistency, and allow the algorithm to recurse again. Suppose now $choose\_literal$ were to pick $not\ p$. Following the same steps outlined earlier, the algorithm would derive inconsistency. Upon backtracking, the algorithm would also derive inconsistency from the selection of $p$. However, the finding would only affect the current branch of the search stemming from the selection of $u(1)$, and the algorithm would then backtrack, select $not\ u(1)$, and recurse. At this point, the algorithm would be again free to select any of the remaining $u(X)$ literals, which from an intuitive point of view means going in the wrong direction. Even if the algorithm were to select $not\ p$ right away, it would still have to backtrack over the choice of $u(0)$ and explore the corresponding branch of the search tree that starts from $not\ u(0)$ before finally concluding that the program is inconsistent. The reader can imagine the effect on the algorithm's performance if *choose_literal* were to choose *not p* at an even later point in the search process.

In order to reduce the chances of *choose_literal* making "wrong" selections, modern solvers base literal selection on carefully designed heuristics. For example, in SMODELS the selection is roughly based on maximizing the number of consequences that can be derived after selecting the given extended literal [9]. These techniques work well in a number of cases, but not always. In fact, particular features of the program can confuse the heuristics. When that happens at an early stage of the search process, the effect is often disastrous, causing the solver to fail to return an answer in an acceptable amount of time. Particularly frustrating is the fact that the efficiency of the heuristics may change largely in correspondence of small elaborations of the program in input. For example, the *choose_literal* heuristics may make good selections for one problem instance, while they may cause the search to take an unacceptable amount of time for a not-too-different problem instance.

As we mentioned in the introduction, one way to limit the effect of wrong selections by *choose_literal* is that of allowing the solver to learn about relevant conflicts at run-time. Once learned, the information about conflicts can be used for the early pruning of other branches of the search space (e.g. [5,4]). Although this technique has proven to be extremely effective, it does not address directly the issue of *choose_literal* making wrong choices, but rather curbs the problem by making some of those choices impossible after learning has taken place, or by allowing to quickly backtrack after a wrong choice has been made. Furthermore, because the learning occurs at run-time, during the initial phase of the computation in which learning has not yet occurred, *choose_literal* may once again affect efficiency negatively by taking the search process in the wrong direction. Finally, whatever has been learned in one execution of the algorithm is discarded upon termination, and cannot be used in later runs.

In the next section, we describe a different approach, aimed at improving directly the selections made by *choose_literal* and at retaining what the algorithm has learned.

## 4 The DORS Framework

Our technique for learning domain-specific heuristics and using them for literal selection applies to the situation in which one is interested in solving a number of problem instances from a given problem domain. Such situations are very common in the ASP community – see e.g. the Second Answer Set Programming Competition [11]. Moreover, this is particularly the case in industrial applications, where the application contains the domain description, and the user describes the instance using some interface (refer e.g. to [12]), which then automatically encodes the problem instance.

Program $P_1$, shown earlier, can be viewed as consisting of a domain description and a problem instance: the first 7 rules constitute the former, while the definition of predicate $t$ is the problem instance. A different problem instance might then define $t$ as $\{t(5), t(6), t(7)\}$. In this case, it is obvious that a good strategy for the selection of the literals consists in first choosing among $\{p, \text{not } p, q, \text{not } q\}$ and only later (if necessary) considering the extended literals formed by $u$ and $v$.

In general, the domain-specific heuristics for *choose_literal* will be learned – rather than manually specified – by analyzing the choices made by the standard solver *solve* when solving representative problem instances from the domain. This approach is particularly useful in applications in which a number of problem instances from the same class of problems will have to be solved over time – for example, in the setting of an industrial application, or in a programming/solver competition in which benchmarking is involved – and computational power is available off-line to allow learning the domain-specific heuristics (e.g. before deploying the application, or before submitting the solver or solutions to a competition).

Next, we discuss how choices made in previous runs of the algorithm can be extracted and combined for future use. The final result will be the learning of a *policy* (see e.g. [13] for a comprehensive introduction on the topic), that is, in general terms, of a mapping from states to probabilities of selecting each available action. To achieve this, the algorithm from Figure 1 is modified to maintain a record of the choice points selected, and to return the list of such choice points together with the answer set, when one is found. The modified algorithm is shown in Figure 2. In the algorithm, the list of

```
function solvecp ( Π : Program, A : Set of Extended Literals, S : Ordered List of Extended Literals )
    B := expand(Π, A);
    if (B is answer set of Π) then return ⟨B, S⟩;
    if (B is not consistent or B is complete) then return ⊥;
    e := choose_literal(Π, B);
    ⟨B′, S′⟩ := solve(Π, B ∪ {e}, S ∘ e);
    if (B′ ≠ ⊥) then return ⟨B′, S′⟩;
    ⟨B′, S′⟩ := solve(Π, B ∪ {not(e)}, S ∘ not(e));
    return ⟨B′, S′⟩;
```

**Figure 2.** Search Algorithm for ASP with Explicit Tracking of Choice Points

choice points is stored in variable $S$. Symbol $\circ$ represents concatenation. When *solvecp* is initially invoked, $S$ is the empty list.

Now we turn our attention to combining the information collected by *solvecp* into a domain-specific heuristics. Given the domain description $M$ and a problem instance $I$ that is to be used to learn the domain-specific heuristics, the *decision-sequence* of $I$ (denoted by $d(I)$) is $\bot$ if $solvecp(I \cup M, \emptyset, \emptyset) = \bot$ and $S$ if $solvecp(I \cup M, \emptyset, \emptyset) = \langle A, S \rangle$ for some $A$. From now on, given a decision-sequence $d$, we denote its $n^{th}$ element by $d_n$. Moreover, given an extended literal $e$, $level(e, d)$ denotes the index $i$ such that $d_i = e$ ($e$ is guaranteed not to occur at more than one position by construction of the decision-sequence in *solvecp*). Intuitively, $level(e, d)$ represents the level in the decision tree at which $e$ was selected. Notice that, by construction of the sequence of choice points in *solvecp*, if $d(I) \neq \bot$, then $d(I)$ only enumerates the choice points that led directly to the answer sets. All the choice points that did not lead directly to it, in the sense that they were later backtracked upon, are in fact discarded every time the algorithm backtracks.

In order to improve the quality of the learned heuristics, we divide the class of problem instances in subclasses, and associate with each problem instance $I$ an expression

$\sigma$ denoting the subclass it belongs to. The intuition is that using subclasses allows to further tailor the literal selection heuristics to the particular features of the problem instances. For example, in a planning domain, $\sigma$ might be the maximum length of the plan (often called $lasttime$ or $maxtime$ in ASP-based planning). The subclass of a problem instance $I$ is denoted by $\sigma(I)$.

Let $\mathcal{I}$ denote the set of all problem instances that will be used for the learning of the domain-specific heuristics. Next, we specify a way to determine how many times an extended literal $e$ was selected at a certain level of the decision-sequences for the problem instances in $\mathcal{I}$. More precisely, given a positive integer $\delta$, called the *scaling factor*, and subclass $\sigma$, the *occurrence count* of an extended literal $e$ w.r.t. a level $l$ and set of instances $\mathcal{I}$ is

$$o_{\delta,\sigma}(e, l, \mathcal{I}) = |\ \{\ I\ |\ I \in \mathcal{I}\ \wedge\ \sigma(I) = \sigma\ \wedge\ d(I) \neq \bot\ \wedge\\ l - \delta/2 \leq index(e, d(I)) < l + \delta/2\ \}\ |.$$

The scaling factor $\delta$ allows taking into account all the occurrences of $e$ at a level in the interval $[l - \delta/2, l + \delta/2)$. If $\delta = 1$, then only the occurrences of $e$ with level equal to $l$ are considered. Values of $\delta$ greater than 1 can be useful in those cases in which all or most permutations of a subsequence of choice points lead to an answer set.

Let now $E = \{e_1, e_2, \ldots, e_k\}$ be a set of extended literals, representing possible choice points at some level $l$ of the decision tree. The *set of best choice points* among $E$ is:

$$best_\delta(l, E, \sigma, \mathcal{I}) = \{e\ |\ e \in E\ \wedge\ \forall e' \in E\ \ o_{\delta,\sigma}(e, l, \mathcal{I}) \geq o_{\delta,\sigma}(e', l, \mathcal{I})\}.$$

Intuitively, $best_\delta(l, E, \sigma, \mathcal{I})$ returns the choice points that, if taken at level $l$, are most likely to lead to an answer set without backtracking, based on the information collected about the instances of subclass $\sigma$ in $\mathcal{I}$. Algorithms for the computation of $best_\delta(l, E, \sigma, \mathcal{I})$ and $o_{\delta,\sigma}(e, l, \mathcal{I})$ are simple and are omitted to save space.

Function $best_\delta(l, E, \sigma, \mathcal{I})$ encodes the essence of the domain-specific heuristics, or, more precisely, the policy[1] for the selection of choice points. Algorithm $choose\_literal$ can now be extended to perform literal selection guided by the domain-specific heuristics. The modified algorithm, $choose\_literal\_dspec$, is shown in Figure 3. In $choose\_literal\_dspec$, argument $T$ is the set of extended literals that have previously been selected by $choose\_literal\_dspec$. If $best_\delta(level, E', \sigma(I), \mathcal{I})$ is the empty set, then $choose\_literal\_dspec$ falls back to performing standard extended literal selection via $choose\_literal$. This is for instances in which the learned heuristics do not prescribe any extended literal for the current decision level, or in which all the extended literals that the learned heuristics prescribed have already been tried. Modifying the standard solver's algorithm in order to use the domain-specific heuristics for choice-point selection is rather straightforward. A simple version, which for the most part follows the well-known iterative version of the SMODELS algorithm, is shown in Figure 4.

Next, we describe how grounding is handled in the DORS framework. The discussion is based on the architecture of the LPARSE+SMODELS system but can be extended to other ASP systems as well. ASP solvers typically expect in input ground

---

[1] We assume uniform probability of selection among the elements of the set returned by $best_\delta(l, E, \sigma, \mathcal{I})$.

```
function choose_literal_dspec ( Π : Program, σ : Problem Subclass, A : Set of Extended Literals,
                                 level : Integer /* Current Level in the Decision Tree */,
                                 T : Set of Extended Literals, I : Set of Instances,
                                 δ : Integer /* Scaling Factor*/ )

    L := lit(Σ(Π));  E := L ∪ {not l | l ∈ L};
    E' = ∅;
    for each e ∈ E
        if  (e ∉ A ∧ not(e) ∉ A ∧ e ∉ T)  then  E' := E' ∪ {e};
    end for
    B := best_δ(level, E', σ, I);
    if  (B ≠ ∅)  then  chosen := one_element_of(B);
                  else  chosen := choose_literal(Π, A);

    return chosen;
```

**Figure 3.** Function for Literal Selection with Domain-Specific Heuristics

(i.e. variable-free) programs. Because however using variables in ASP programs is convenient, programs are first pre-processed by a *grounder* (LPARSE and GRINGO in the systems considered here), which replaces each non-ground rule by the set of its ground instances. The main difficulty in implementing our technique in state-of-the-art ASP systems is that their grounders often introduce "unnamed atoms" during the grounding process. An unnamed atom is an atom that does not occur in the original program, and is used internally by the ASP system. Because of their local use, unnamed atoms are assigned identifiers that are only valid for the current run of the system. There is no guarantee that unnamed atoms will be assigned the same identifiers when the system is run on a different problem instance. Because nothing prevents unnamed atoms from being used as choice points by the solver, one needs to ensure that unnamed atoms are given a unique, known identifier, so that choice-point information regarding them can be properly handled. One possible solution is to modify the ASP grounders so that unnamed atoms are given identifiers that remain valid across multiple executions. Although conceptually simple, this solution requires modifying each grounder that one is interested in using. In this paper we present instead a relatively simple, indirect method that consists of a pre-processing phase and a post-processing phase, and does not involve modifications to the grounders.

In LPARSE and GRINGO, unnamed atoms are introduced during the grounding of rules containing certain constraint literals, in order to simplify their structure.[2] For example, the choice rule in the program:

$$\begin{cases} p(1).\, p(2).\, p(3).\\ 1\{a(X) : p(X)\}2. \end{cases}$$

is translated by the grounder as:

$$\begin{cases} \{a(1), a(2), a(3)\}.\\ \leftarrow \mu_1.\\ \mu_1 \leftarrow 3\{\text{not } a(1), \text{not } a(2), \text{not } a(3)\}.\\ \leftarrow \mu_0.\\ \mu_0 \leftarrow 3\{a(3), a(2), a(1)\}. \end{cases}$$

---

[2] A thorough explanation of the process is beyond the scope of this paper. We refer the interested reader to e.g. [8].

```
function solve_dspec ( Π : Program, σ : Problem Subclass, I : Set of Instances, δ : Scaling Factor )
  var S : Stack of Sets of Extended Literals;
  var B, T : Set of Extended Literals;
  var terminate : Boolean;

    S := ∅;  B := ∅;  T := ∅;
    terminate := false;
    while (terminate = false)
        B := expand(Π, B);
        if (B is answer set of Π) then
            terminate := true;
        else
            if (B is not consistent or B is complete) then
                if (S = ∅) then
                    B := ⊥;
                    terminate := true;
                else
                    /* Backtrack */
                    B := top(S);
                    S := pop(S);
                end if
            else
                /* Select a choice point */
                e := choose_literal_dspec(Π, σ, B, level, T, I, δ);
                T := T ∪ {e};
                S := push(B ∪ {not(e)}, S);
                B := B ∪ {e};
            end if
        end if
    end while
    return B;
```

**Figure 4.** Search Algorithm for ASP with Domain-Specific Heuristics for Choice-Point Selection

where $\mu_0$ and $\mu_1$ are unnamed atoms. As we mentioned above, no assumptions can be made about which identifiers are used for the unnamed atoms. If we were for example to add to the program a second choice rule[3], the grounding of the new program could use some new identifiers $\mu_2$, $\mu_3$ for the above translation. On the other hand, because of the structure of the grounding algorithm, *the relative order of the rules belonging to the grounding of the choice rule is independent from the changes made to the rest of the program.* Moreover, whenever multiple unnamed atoms occur in the body of a rule, *their relative order is independent of changes made to the rest of the program.* We will make use of these two properties later.

In the pre-processing phase, the user specifies a name for each rule whose grounding may cause the introduction of unnamed atoms. Because we want to avoid modifications to the grounder, we cannot extend the syntax of rules to allow specifying a name explicitly.[4] Therefore, the name of the rule is rather specified in the body of the rule, using a

---

[3] Or if the number of ground instances of the choice rule of our example were to change because of changes in the problem instance.

[4] However, a pre-processor can be used, as discussed later.

special relation $\nu$.[5] So, the choice rule above can be written as:

$$1\{a(X) : p(X)\}2 \leftarrow \nu(r_1). \tag{1}$$

Generally speaking, given a list, $X$, of all the free variables in the rule, and some fresh constant $\rho$, the name is specified by the atom $\nu(\rho, X)$. A rule whose name is specified as above is called an *augmented* rule.

To ensure that the meaning of a rule is not altered by the augmentation, a definition of atom $\nu(\cdot)$ must also be provided (otherwise the body of the augmented rule is never satisfied). Because state-of-the-art grounders usually drop trivially-true atoms from the body of the rules, we define the new atom by a choice rule with no bounds and suitable domain predicates for the arguments of relation $\nu$, such as $\{\ \nu(r_1)\ \}$. (The choice rule will be removed later, to avoid affecting the performance of the solver.) When processing (1), the grounder produces:

$$\begin{cases} \{a(1), a(2), a(3)\} \leftarrow \nu(r_1). \\ \leftarrow \mu_1, \nu(r_1). \\ \mu_1 \leftarrow 3\{\text{not } a(1), \text{not } a(2), \text{not } a(3)\}. \\ \leftarrow \mu_0, \nu(r_1). \\ \mu_0 \leftarrow 3\{a(3), a(2), a(1)\}. \end{cases}$$

Notice how the unnamed atoms co-occur with the $\nu(\cdot)$ atom in the body of some of the rules. Because of the structure of the grounding algorithm, *this is the case for the grounding of any rule that introduces unnamed atoms.* The reader should also notice that the addition of $\nu(\cdot)$ atoms to the program can be easily automated. A user could then specify a name for the rule using a more convenient syntax, and have a simple pre-processor introduce the $\nu(\cdot)$ atoms in the program as shown above.

The post-processing phase is based on the algorithm shown in Figure 5. The al-

```
function postp ( G : GroundProgram )
    Assoc := ∅;  G' := G;
    for each rule ρ ∈ G and unnamed atom μ in ρ
        if ρ contains an atom ν(X) for some X then
            i := smallest positive integer such that ∀μ' ⟨μ', ν(i, X)⟩ ∉ Assoc;
            Assoc := Assoc ∪ {⟨μ, ν(i, X)⟩};
        end if
    end for
    for each atom of the form ν(X)
        Remove from G' rule {ν(X)} ← Γ (for some Γ);
        Remove every occurrence of ν(X) from G';
    end for
    for each ⟨μ, ν(i, X)⟩ ∈ Assoc
        Replace all occurrences of μ in G' by ν(i, X);
    end for
    return G';
```

**Figure 5.** Post-processing algorithm

gorithm works as follows. First, the ground rules are scanned for co-occurrences of unnamed atoms and $\nu$ atoms. The goal is to use the information provided by the $\nu$ atoms to give a name to the unnamed atoms they co-occur with. The association of names to unnamed atoms is stored in variable $Assoc$. Because multiple unnamed atoms may be introduced by the grounding of a single rule, an extra integer argument is added to relation $\nu$ when naming unnamed atoms. Values for that argument are assigned on a first-come, first-serve basis. Because, as we noted above, the relative order of unnamed atoms in the ground rules does not change, we are guaranteed that the naming of unnamed atoms will be consistent throughout multiple runs of the grounder with different input programs (as long as the domain description remains the same). In the next *for* loop, all $\nu$ atoms and their definitions are removed from the program. Finally, the unnamed atoms are renamed according to the associations encoded by variable $Assoc$.

---

[5] Notice that the specification of the name of the rule in the body is purely a technical device, and should not be intended to convey any semantic information.

## 5 Experimental Evaluation

In this section we discuss an experimental evaluation of the DORS framework. To ensure applicability to a wider variety of cases, we have tested our implementation on both abstract problems and on problems from industrial applications of ASP. Here we show the results of testing on the 15 puzzle problem and on the task of planning for the Reaction Control System of the Space Shuttle.

The solver used in the experiments is SMODELS, which we modified to obtain implementations of algorithms $solvecp$ and $solve\_dspec$. It should be noted that we did not use CLASP for our experiments. In fact, extending the DORS framework to CLASP is complicated by the fact that this solver is based on conflict-driven clause learning (CDCL) (e.g. [5]) rather than DPLL. Although we believe that certain similarities between DPLL and CDCL make it technically possible to extend the DORS framework to CDCL-based systems, work on implementing the DORS framework within CLASP is still in the early stages and results will be discussed in a later paper.

In the rest of the discussion, we refer to the implementation of $solve\_dspec$ within SMODELS as DSPEC. The grounders used were GRINGO for the 15 puzzle (because the original solution of the puzzle used some features specific of GRINGO's language) and LPARSE for the Reaction Control System. *It is important to note that this interchangeable use of grounders is only possible because of the grounding technique we described in the previous section.*

The 15 puzzle problem was one of the benchmarks used for the Second ASP Programming Competition [11]. The description of the puzzle, taken from the competition's web site, is shown in Figure 6a. The goal configuration used in the competition is shown in Figure 6b. For the domain description, we have used the program published on the competition's web site [6], modified to provide names of select rules, as explained earlier. Next, for every value of $k$ ranging between 10 and 30, we have generated 100 random problem instances that can be solved with $k$ moves or less. The subclass that a problem instance belongs to is identified by the value of $k$ (i.e. the maximum allowed length of a plan for that instance, called $maxtime$ in the original encoding). Next, we ran all the instances in each subclass with a timeout value of 6000 sec. The instances

---

[6] The collection is available from http://www.cs.kuleuven.be/~dtai/events/ASP-competition/problem_instances.tar.gz.

*"In 15-Puzzle, we have a $4 \times 4$ grid where there are 15 numbers (1 to 15) and one blank. The goal is to arrange the numbers from their initial configuration to the goal configuration by swapping one number at a time with its adjacent blank position. Let $(x, y)$ be the coordinates of a number on the grid and $(i, j)$ be those of the blank. Then $(x, y)$ and $(i, j)$ are adjacent, if $|x - i| + |y - j| = 1$."*

|   | 1 | 2 | 3 | 4 |
|---|---|---|---|---|
| 1 | 0 | 1 | 2 | 3 |
| 2 | 4 | 5 | 6 | 7 |
| 3 | 8 | 9 | 10 | 11 |
| 4 | 12 | 13 | 14 | 15 |

x/y

**Figure 6.** (a) Description of the 15 Puzzle; (b) Goal Configuration

that took more than a time threshold $t_k$ were then used to learn the domain-specific heuristics, while the remaining instances – called hard instances – were used for the evaluation phase. In the evaluation phase, we have run the hard instances using the learned domain-specific heuristics. The scaling factor $\delta$ (discussed earlier) was set to 1. Figure 7 compares the performance of SMODELS and DSPEC for the subclass with $maxtime = 28$, where we set the threshold $t_k$ to 70 seconds (selected to have a sufficient number of samples for learning). The domain-specific heuristics gave an average speedup of 6.4 times[7] over the standard solver, with a maximum speedup of more than 24. What's more important, out of 11 instances for which the standard solver timed out, all were solved within the time limit by DSPEC, substantiating our claim that the use of domain-specific heuristics helps to make solver's performance more consistent. (Similar performance was obtained on other subclasses. We omit the results to save space.)

| Instance | SMODELS (sec) | DSPEC (sec) | Speedup (times) | Instance | SMODELS (sec) | DSPEC (sec) | Speedup (times) |
|---|---|---|---|---|---|---|---|
| 38 | 95.754 | 3.859 | 24.8 | 85 | 176.233 | 138.327 | 1.3 |
| 40 | 72.831 | 3.714 | 19.6 | 86 | 6000 | 758.293 | 7.9 |
| 71 | 6000 | 959.157 | 6.3 | 87 | 397.587 | 173.338 | 2.3 |
| 73 | 6000 | 600.448 | 10.0 | 88 | 6000 | 706.968 | 8.5 |
| 74 | 1226.096 | 654.586 | 1.9 | 89 | 402.882 | 178.329 | 2.3 |
| 75 | 317.459 | 156.184 | 2.0 | 90 | 267.338 | 173.345 | 1.5 |
| 76 | 6000 | 987.010 | 6.1 | 91 | 439.324 | 247.809 | 1.8 |
| 77 | 415.760 | 129.066 | 3.2 | 92 | 391.290 | 175.358 | 2.2 |
| 78 | 427.494 | 170.919 | 2.5 | 93 | 6000 | 769.548 | 7.8 |
| 79 | 6000 | 780.723 | 7.7 | 94 | 6000 | 830.665 | 7.2 |
| 80 | 270.246 | 167.075 | 1.6 | 95 | 314.586 | 178.609 | 1.8 |
| 81 | 6000 | 473.997 | 12.7 | 96 | 6000 | 865.279 | 6.9 |
| 82 | 183.667 | 162.769 | 1.1 | 98 | 1100.905 | 72.243 | 15.2 |
| 83 | 471.565 | 137.465 | 3.4 | 99 | 725.745 | 50.194 | 14.5 |
| 84 | 6000 | 845.182 | 7.1 | 100 | 661.419 | 74.023 | 8.9 |

**Figure 7.** Performance Comparison on the 15 Puzzle. Machine specs: Intel Q6600 CPU, 2.4GHz, 6GB RAM.

---

[7] This is just the lower bound of the estimate, since SMODELS timed out several times.

The second problem domain for which we report experimental results is that of planning for the Reaction Control System (RCS) of the Space Shuttle. As described in e.g. [14,12], the RCS is the Shuttle's system that has primary responsibility for maneuvering the Shuttle while it is in space. It consists of fuel and oxidizer tanks, valves, and other plumbing needed to provide propellant to the maneuvering jets of the Shuttle. The RCS also includes electronic circuitry, both to control the valves in the fuel lines and to prepare the jets to receive firing commands. In order to configure the Shuttle for an orbital maneuver, the RCS must be configured by opening and closing appropriate valves. This is accomplished by either changing the position of the associated switches, or by issuing computer commands. In normal conditions, the procedures for the configuration of the RCS for a given maneuver are known in advance by the astronauts. However, if components of the RCS are faulty, then the standard procedures may not be applicable. Moreover, because of the amount of possible combinations of faults, it is impossible to prepare in advance a set of configuration procedures for faulty situations. In those cases, ground control needs to carefully examine the problem and manually come up with a configuration procedure. The system described in [14,12] uses a model of the RCS, as well as ASP-based reasoning algorithms, to provide ground control with a decision-support system that automatically generates configuration procedures for the RCS and that can be used when faulty components are present (incidentally, the system can also perform diagnostic reasoning [12]).

A collection of problem instances from the domain of the RCS is publicly available, together with the ASP encoding of the model of the RCS.[8] The interested reader may refer to [14] for a description of the instances. For our testing, we have selected a set of 425 instances from the collection, corresponding to the public instances with no electrical faults and 3, 8, and 10 mechanical faults respectively, for which a plan of length 6 or less (determined by parameter $lasttime$) was found in the experiments discussed in [14,12], and we have analyzed the performance of the solver on planning with maximum lengths ranging between 6 and 10.

As before, first we ran all the instances with the standard solver and a timeout of 6000 sec. Of those, the instances that took less than 50 sec were used to learn the domain-specific heuristics, while the remaining "hard instances" were used for the evaluation phase. The problem subclasses were defined by the pair $\langle lasttime, maneuver \rangle$, where $lasttime$ specifies the maximum plan length and $maneuver$ is the maneuver that the RCS must be configured for (in our experiments, using the maneuver in the subclass definition substantially improved the performance of the learned heuristics). Figure 8 shows the results of the comparison for the 91 hard instances with 8 and 10 mechanical faults and values of lasttime of 9 and 10. We believe the speedup obtained with the domain-specific heuristics is remarkable. First of all, out of 53 instances for which the standard solver timed out before finding a solution, in 48 cases the domain-specific heuristics allowed to find a solution within the time limit, and in some cases in under 10 seconds. The average speedup is 259.2, with a peak of 1253.1 for an in-

---

[8] The files are available from http://www.krlab.cs.ttu.edu/Software/Download/.

**8 Mechanical Faults**

| Lasttime/Instance | SMODELS (sec) | DSPEC (sec) | Speedup (times) | Lasttime/Instance | SMODELS (sec) | DSPEC (sec) | Speedup (times) | Lasttime/Instance | SMODELS (sec) | DSPEC (sec) | Speedup (times) |
|---|---|---|---|---|---|---|---|---|---|---|---|
| 9 / 025 | 6000 | 17.643 | 340.1 | 9 / 191 | 4829.019 | 8.869 | 544.5 | 10 / 096 | 789.351 | 13.787 | 57.3 |
| 9 / 027 | 6000 | 9.597 | 625.2 | 9 / 199 | 437.379 | 7.144 | 61.2 | 10 / 103 | 6000 | 16.781 | 357.5 |
| 9 / 038 | 125.244 | 8.616 | 14.5 | 10 / 013 | 94.623 | 21.663 | 4.4 | 10 / 110 | 6000 | 255.421 | 23.5 |
| 9 / 044 | 1439.027 | 6.846 | 210.2 | 10 / 022 | 6000 | 423.565 | 14.2 | 10 / 113 | 264.419 | 6000 | 0.044 |
| 9 / 053 | 6000 | 13.599 | 441.2 | 10 / 025 | 6000 | 2035.089 | 2.9 | 10 / 120 | 1983.466 | 20.254 | 97.9 |
| 9 / 059 | 85.151 | 551.806 | 0.2 | 10 / 027 | 6000 | 10.248 | 585.5 | 10 / 140 | 64.451 | 6000 | 0.011 |
| 9 / 074 | 6000 | 8.961 | 669.6 | 10 / 032 | 2949.169 | 13.820 | 213.4 | 10 / 147 | 187.800 | 7.125 | 26.4 |
| 9 / 075 | 736.134 | 3.837 | 191.9 | 10 / 037 | 6000 | 12.218 | 491.1 | 10 / 154 | 942.008 | 6000 | 0.157 |
| 9 / 087 | 6000 | 6000 | 1.0 | 10 / 044 | 6000 | 18.162 | 330.4 | 10 / 165 | 6000 | 30.008 | 199.9 |
| 9 / 090 | 6000 | 14.111 | 425.2 | 10 / 050 | 72.596 | 12.521 | 5.8 | 10 / 166 | 6000 | 820.789 | 7.3 |
| 9 / 093 | 2451.649 | 6.477 | 378.5 | 10 / 053 | 1907.445 | 23.370 | 81.6 | 10 / 177 | 6000 | 12.605 | 476.0 |
| 9 / 098 | 114.643 | 10.529 | 10.9 | 10 / 059 | 6000 | 15.163 | 395.7 | 10 / 178 | 6000 | 6000 | 1.0 |
| 9 / 103 | 52.219 | 12.544 | 4.2 | 10 / 061 | 266.024 | 7.756 | 34.3 | 10 / 179 | 6000 | 16.740 | 358.4 |
| 9 / 122 | 6000 | 4.788 | 1253.1 | 10 / 070 | 519.583 | 16.343 | 31.8 | 10 / 188 | 5235.985 | 12.740 | 411.0 |
| 9 / 140 | 6000 | 11.493 | 522.1 | 10 / 074 | 6000 | 13.903 | 431.6 | 10 / 189 | 3773.981 | 11.765 | 320.8 |
| 9 / 165 | 6000 | 13.027 | 460.6 | 10 / 077 | 251.754 | 7.518 | 33.5 | 10 / 190 | 6000 | 1010.510 | 5.9 |
| 9 / 170 | 6000 | 6000 | 1.0 | 10 / 087 | 6000 | 24.962 | 240.4 | 10 / 194 | 6000 | 12.407 | 483.6 |
| 9 / 179 | 6000 | 14.304 | 419.5 | 10 / 088 | 3830.141 | 18.512 | 206.9 | 10 / 199 | 6000 | 9.452 | 634.8 |
| 9 / 184 | 6000 | 20.254 | 296.2 | 10 / 092 | 318.830 | 11.712 | 27.2 | | | | |
| 9 / 188 | 6000 | 6000 | 1.0 | 10 / 093 | 6000 | 494.850 | 12.1 | | | | |

**10 Mechanical Faults**

| Lasttime/Instance | SMODELS (sec) | DSPEC (sec) | Speedup (times) | Lasttime/Instance | SMODELS (sec) | DSPEC (sec) | Speedup (times) | Lasttime/Instance | SMODELS (sec) | DSPEC (sec) | Speedup (times) |
|---|---|---|---|---|---|---|---|---|---|---|---|
| 9 / 011 | 6000 | 12.745 | 470.8 | 10 / 014 | 6000 | 15.671 | 382.9 | 10 / 081 | 6000 | 23.573 | 254.5 |
| 9 / 036 | 223.258 | 5.546 | 40.3 | 10 / 035 | 6000 | 18.642 | 321.9 | 10 / 082 | 6000 | 24.662 | 243.3 |
| 9 / 062 | 6000 | 14.127 | 424.7 | 10 / 036 | 327.915 | 8.000 | 41.0 | 10 / 094 | 6000 | 16.976 | 353.4 |
| 9 / 077 | 6000 | 7.663 | 783.0 | 10 / 052 | 4878.440 | 13.202 | 369.5 | 10 / 096 | 6000 | 19.369 | 309.8 |
| 9 / 082 | 371.375 | 17.191 | 21.6 | 10 / 053 | 6000 | 11.523 | 520.7 | 10 / 099 | 6000 | 20.283 | 295.8 |
| 9 / 083 | 6000 | 11.068 | 542.1 | 10 / 056 | 5585.635 | 11.041 | 505.9 | 10 / 115 | 6000 | 14.852 | 404.0 |
| 9 / 096 | 550.659 | 16.294 | 33.8 | 10 / 059 | 6000 | 6000 | 1.0 | 10 / 133 | 6000 | 5.812 | 1032.3 |
| 9 / 104 | 6000 | 13.471 | 445.4 | 10 / 062 | 6000 | 19.492 | 307.8 | 10 / 136 | 3144.372 | 14.080 | 223.3 |
| 9 / 115 | 6000 | 2141.781 | 2.8 | 10 / 064 | 6000 | 14.095 | 425.7 | 10 / 143 | 67.685 | 6000 | 0.011 |
| 9 / 175 | 79.346 | 9.756 | 8.1 | 10 / 072 | 556.285 | 6000 | 0.1 | 10 / 147 | 6000 | 5.110 | 1174.2 |
| 10 / 008 | 172.772 | 19.576 | 8.8 | 10 / 078 | 748.273 | 5.797 | 129.1 | 10 / 180 | 6000 | 170.726 | 35.1 |

**Figure 8.** Performance Comparison on the RCS Domain. Machine specs: Intel i7 CPU, 2.93GHz, 8GB RAM.

stance for which SMODELS timed out[9], and a peak of $544.5$ for an instance for which SMODELS did not time out. In $6$ cases (out of $91$) DSPEC performed worse than the standard solver. We believe that these outliers can be eliminated if more samples are made available for learning.

## 6 Conclusions

In this paper we have described a framework that allows learning and using domain-specific heuristics for choice-point selection, and we have demonstrated its application to ASP. Our experimental evaluation has shown that domain-specific heuristics can give remarkable speedups, and allow to find answers that cannot otherwise be computed in a reasonable amount of time. In the case of the RCS domain, a large number of the instances for which the standard solver timed out, could be solver in a matter of seconds using the domain-specific heuristics, with an average speedup of more than 2 orders of magnitude and peaks of more than 3. This is the type of consistent performance that makes a solver viable for industrial applications. We believe that an appealing feature of the DORS framework is that in principle it can be applied to any solver based on the DPLL procedure. Hence, it is possible to extend the approach shown here to other ASP solvers, or even to e.g. constraint solvers. Work is also ongoing on extending the DORS framework to solvers based on conflict-driven clause learning, such as CLASP. As a final note, we would like to point out that the method used here to learn the domain-specific heuristics is a very simple instance of policy learning. It will be interesting to investigate how more sophisticated techniques from reinforcement learning, but also from machine learning and data mining, can be applied within the DORS framework. We expect that doing so will allow to improve performance of the solvers even further.

---

[9] The actual speedup could in fact be higher, since SMODELS timed out. As a test, we have let SMODELS run on some of these instances for over $60,000$ seconds (16 hours) without getting a solution.

## References

1. Davis, M., Putnam, H.: A Computing Procedure for Quantification Theory. Communications of the ACM **7** (1960) 201–215
2. Davis, M., Logemann, G., Loveland, D.: A Machine program for theorem proving. Communications of the ACM **5**(7) (1962) 394–397
3. Gebser, M., Kaufmann, B., Schaub, T.: The Conflict-Driven Answer Set Solver clasp: Progress Report. In: 10th International Conference on Logic Programming and Nonmonotonic Reasoning (LPNMR09). (Sep 2009) 509–514
4. Gebser, M., Kaufmann, B., Neumann, A., Schaub, T.: Conflict-driven answer set solving. In Veloso, M.M., ed.: Proceedings of the Twentieth International Joint Conference on Artificial Intelligence (IJCAI'07), MIT Press (2007) 386–392
5. Goldberg, E., Novikov, Y.: BerkMin: A Fast and Robust Sat-Solver. In: Proceedings of Design, Automation and Test in Europe Conference (DATE-2002). (Mar 2002) 142–149
6. Gelfond, M., Lifschitz, V.: Classical negation in logic programs and disjunctive databases. New Generation Computing **9** (1991) 365–385
7. Marek, V.W., Truszczynski, M.: Stable models and an alternative logic programming paradigm. In: The Logic Programming Paradigm: a 25-Year Perspective. Springer Verlag, Berlin (1999) 375–398
8. Niemela, I., Simons, P.: Extending the Smodels System with Cardinality and Weight Constraints. In: Logic-Based Artificial Intelligence. Kluwer Academic Publishers (2000) 491–521
9. Niemela, I., Simons, P., Soininen, T.: Extending and implementing the stable model semantics. Artificial Intelligence **138**(1–2) (Jun 2002) 181–234
10. Leone, N., Pfeifer, G., Faber, W., Eiter, T., Gottlob, G., Perri, S., Scarcello, F.: The DLV System for Knowledge Representation and Reasoning. ACM Transactions on Computational Logic **7**(3) (2006) 499–562
11. Denecker, M., Vennekens, J., Bond, S., Gebser, M., Truszczynski, M.: The Second Answer Set Programming Competition. In: 10th International Conference on Logic Programming and Nonmonotonic Reasoning (LPNMR09). (Sep 2009) 637–654

12. Balduccini, M., Gelfond, M., Nogueira, M.: Answer Set Based Design of Knowledge Systems. Annals of Mathematics and Artificial Intelligence (2006)
13. Barto, A.G., Sutton, R.S.: Reinforcement learning: an introduction. MIT Press (1998)
14. Nogueira, M.: Building Knowledge Systems in A-Prolog. PhD thesis, University of Texas at El Paso (May 2003)